\documentclass[final]{cvpr}
\UseRawInputEncoding
\usepackage{times}
\usepackage{epsfig}
\usepackage{graphicx}
\usepackage{amsmath}
\usepackage{amssymb}
\usepackage{subfigure}
\usepackage{multirow} 
\usepackage{xcolor}
\usepackage{booktabs}

\usepackage[pagebackref=true,breaklinks=true,colorlinks,bookmarks=false]{hyperref}

\def\cvprPaperID{78} 
\def\confYear{CVPR 2021}

\begin{document}
	
\title{VSpSR: Explorable Super-Resolution via Variational Sparse Representation}

\author{Hangqi Zhou \quad Chao Huang \quad Shangqi Gao \quad  Xiahai Zhuang\footnote{Xiahai Zhuang is corresponding author.}\\
	School of Data Science, Fudan University\\
	Shanghai, China\\
	{\tt\small https://zmiclab.github.io/}
}

\maketitle
\pagestyle{empty}
\thispagestyle{empty}

\begin{abstract}
	Super-resolution (SR) is an ill-posed problem, which means that infinitely many high-resolution (HR) images can be degraded to the same low-resolution (LR) image. To study the one-to-many stochastic SR mapping, we implicitly represent the non-local self-similarity of natural images and develop a \textbf{V}ariational \textbf{Sp}arse framework for \textbf{S}uper-\textbf{R}esolution (VSpSR) via neural networks. Since every small patch of a HR image can be well approximated by the sparse representation of atoms in an over-complete dictionary, we design a two-branch module, i.e., VSpM,  to explore the SR space. Concretely, one branch of VSpM extracts patch-level basis from the LR input, and the other branch infers pixel-wise variational distributions with respect to the sparse coefficients. By repeatedly sampling coefficients, we could obtain infinite sparse representations, and thus generate diverse HR images. According to the preliminary results of NTIRE 2021 challenge on learning SR space, our team ranks 7-th in terms of released scores. 
	
\end{abstract}

\section{Introduction}
Single image super-resolution (SR) aims at estimating the mapping from low-resolution (LR) to high-resolution (HR) spaces \cite{super02,low01,joint03}. Taking into account that HR images will lose many details in the high-to-low degradation process, the SR problem is naturally underdetermined, which brings the fact that there exists multiple HR images correspond to one input LR image. Although these HR images may have the same low-frequency information in LR space, their high-frequency information, including textures and details, can be significantly different. This ill-posed nature makes SR task a challenging problem to solve. 

Recently, the learning-based approaches have made great progress due to its robust ability of recovering details \cite{real01,accelerate01,deep03,real02,image02,deep02}. When early deep learning methods focus on promoting computational metrics like PSNR and SSIM \cite{wide01,accurate01, enhanced01}, methods proposed later pay more attention to the progress of SR application in real-world. CARN \cite{2018Fast} proposed a lightweight network to speed up training and inference, and Meta-SR \cite{2020Meta} develops a up-sampling module capable of dealing with arbitrary scale factor. In 2017, SRGAN \cite{photo01} introduces adversarial training strategy into super-resolution, since then, many  GAN based SR methods are aiming to obtain SR images with better perceptual quality \cite{2018ESRGAN}. However, these SR methods only use LR-HR image pairs to approximate a deterministic mapping, thus ignoring the ill-pose nature of SR problems.  

The development of SR methods from deterministic mapping to stochastic mapping lies on the transformation from fitting single HR output to fitting the conditional distribution of HR images given the LR input. In order to explore the relationship between low-resolution images and the corresponding diverse high-resolution images,  recently published stochastic super-resolution methods \cite{lugmayr2020srflow,BuhlerRT20DeepSEE,BahatM20Explorable,VarSR} reformulated a challenging goal of learning the conditional distribution. Taking into account that HR images share the same low-frequency information, current stochastic SR methods introduce additional latent variable to affect the high-frequency information of HR image \cite{lugmayr2020srflow,BuhlerRT20DeepSEE,BahatM20Explorable,VarSR}, therefore, by sampling different latent variable, these methods can generate diverse HR images with interpretability.

The NTIRE 2021 \footnote{https://data.vision.ee.ethz.ch/cvl/ntire21/} raised one challenge of learning the super-resolution space. The difficulty in this challenge is from three aspect. First, each individual SR prediction should reach high perceptual quality. Second, the proposed method should be able to sample an arbitrary number of SR images and fully explore the uncertainty induced by the ill-posed nature. Moreover, each individual SR prediction should be consistent with  in the LR space, which restricts the performance of many GAN based SR methods.

In this work, we develop a \textbf{V}ariational \textbf{Sp}arse framework for \textbf{S}uper-\textbf{R}esolution (VSpSR) via neural networks to solve the problems in NTIRE 2021 challenge on learning the super-resolution space. Overall, we assume that the HR image contains the deterministic part and the stochastic  part. As for the deterministic part, we can use any deterministic SR method to obtain from the LR input. For the explorable part, as it has been widely utilized in traditional SR methods, we apply the sparse representation into deep-learning, using the diversity of image representation coefficients to control the diversity of HR image. 

Specifically, we design a two-branch module named VSpM to capture the stochastic mapping of details in HR images. Using the LR image as the input, the basis branch of VSpM outputs patch-level basis in SR space, and the coefficients branch infers pixel-wise variational distributions with respect to the sparse coefficients. Therefore, by repeatedly sampling coefficients, we could obtain infinite sparse representations, and thus generate diverse HR images. Our experiments show that the variational sparse framework leads to larger SR space, and the VSpM module has the potential to cooperate with other deterministic SR methods to enhance their exploration ability. Our methods ranked 7-th in the NTIRE 2021 challenge on learning the super-resolution space according to the preliminary results \cite{lugmayr2021ntire}.

The rest of this work is organized as follows. Section \ref{sec:2} shows the related works about the deterministic and stochastic SR. We develop a variational sparse framework for explorable super-resolution (VSpSR) via neural networks and gives the details of training strategies in Section \ref{sec:3}. The description of our experiments is presented in Section \ref{sec:4}. We discuss the proposed method in Section \ref{sec:5} and conclude this work in Section \ref{sec:6}.

\section{Related works}\label{sec:2}
\textbf{Single image SR:}  SR problem is naturally underdetermined due to the information loss in the high-to-low degradation process. Many traditional SR methods have noticed this fundamental
fact, but they tend to further regularize the problem and finally output \textbf{one} SR prediction \cite{2008Image,2014Anchored}. Recently, DNN has been widely applied in image SR due to its ability of simulating complex mappings. Dong \etal \cite{image02} first proposed to approximate the mapping from LR to HR image pairs using a three layers convolutional neural network. Since then, other architectures, such as RNN \cite{conv01, under01, recurrent01}, ResNet \cite{deep04,2018Fast}, and GAN \cite{gan01,2018ESRGAN}, have been applied in image SR. 
However, previous deep-learning super-resolution methods often use the deterministic mapping $x = f_{\theta}(y)$ to model the process of recovering the high-resolution image $x$ from a given low-resolution image $y$ \cite{enhanced01,2018RCAN}, neglecting the ill-posed nature of this problem. 

\textbf{Stochastic SR:} In order to explore the relationship between low-resolution images and the corresponding diverse high-resolution images,  recently published stochastic super-resolution methods \cite{lugmayr2020srflow,BuhlerRT20DeepSEE,BahatM20Explorable,VarSR} reformulated a challenging goal of learning the conditional distribution $p_{x|y}(x|y, \theta)$. Considering that the low-frequency information of the generated HR image is consistent with that of LR image, current stochastic SR methods tend to introduce additional latent variable $z$ to affect the high-frequency information of HR image to achieve diversity, $x = f_{\theta}(y, z)$ \cite{lugmayr2020srflow,BuhlerRT20DeepSEE,BahatM20Explorable,VarSR}. In this formulation, the exploration of latent variable $z$ controls the diversity of HR image $x$, even the mapping $f_{\theta}$ itself can be deterministic. These stochastic SR methods adopt different strategy to enlarge the SR space. SRFlow \cite{lugmayr2020srflow} adopts the framework of the conditional normalizing flows, using invertible network to restrict the distribution of the latent variable. Yuval Bahat \etal \cite{BahatM20Explorable} propose structure loss and map loss to enhance the affect of control signal, and DeepSEE \cite{BuhlerRT20DeepSEE} generates the latent variable from the semantic information of other high-resolution face images, and thus provides guidance for the generated HR image.

Motivated by the idea of Conditional Variational AutoEncoder (CVAE) \cite{2015Learning} and sparse representation \cite{2008Image}, we propose a variational sparse representation framework, and its details is presented in Section \ref{sec:3}.

\begin{figure}[t]
	\begin{center}
		\includegraphics[width=1\linewidth]{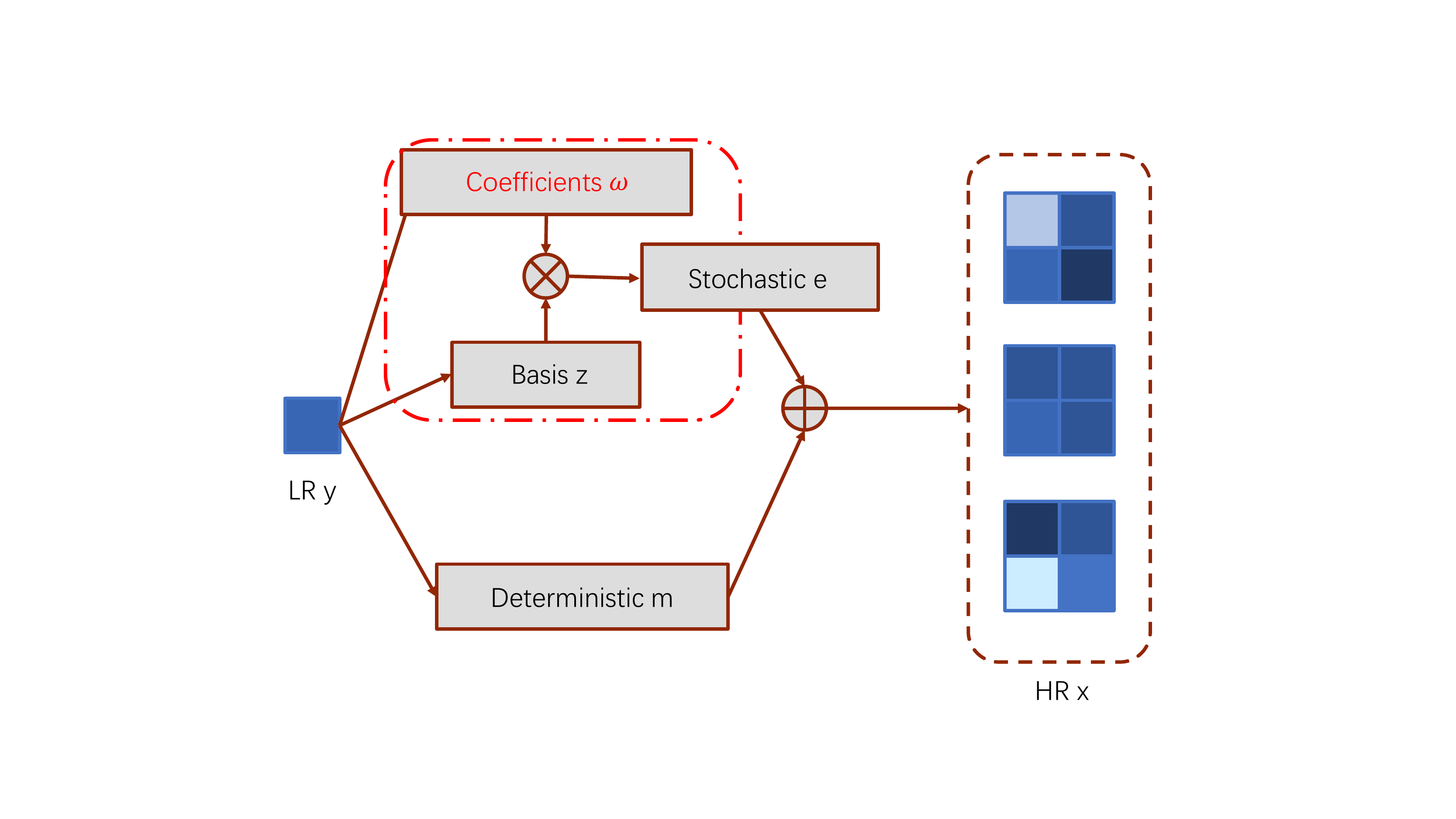}
	\end{center}
	\caption{\textbf{Framework of VSpSR}. We assume that $x$ follows a Gaussian distribution, $x \sim\mathcal N(m+e, \sigma^2I_{d_x})$. (1) We use a deterministic mapping $h$ to get the deterministic part $m$. (2) Through spare representation, the exploration of coefficients $\omega$ controls the diversity of $e$, which can reduce the difficulty in training.}
	\label{fig:bnsr}
\end{figure}

\section{Methodology}\label{sec:3}
Exploring SR space is significant since multiple HR images could be degraded to the same LR images.
However, much attention has been paid to estimate the deterministic mapping from LR to HR images, while few works have been done to explore SR space. 
To rectify the weakness, we proposed a variational sparse representation framework, i.e., VSpSR, as Figure \ref{fig:bnsr} shows to estimate stochastic mappings from single LR image to multiple SR images.
Concretely, we first assume that a HR patch could decomposed into the sum of a low-frequency part $ m $ and a high-frequency component $ e $, where, $ m $ could be deterministic given a LR patch $ y $, but $ e $ is often stochastic. 
Then, we sparsely represent $ e $ via a set of deterministic basis $ z $ and a group of stochastic coefficients $ \omega $.
Moreover, we give a sparse prior to $ \omega $, and infer the distribution of $ \omega $ from $ y $ via variational Bayesian inference. 
Finally, we repeatedly sample the sparse coefficients from the variational distribution, and thus could generate diverse SR patches. 
The statistical model of VSpSR is described in Section \ref{sec:model}, and the detailed network architecture of VSpSR is presented in Section \ref{sec:arch}.

\subsection{Variational sparse representation}\label{sec:model}

Inspired by CVAE \cite{2015Learning} , our method aims to extract latent variables  representing the parameters of the corresponding HR image distribution from the LR image $y$ itself first, and then sample super-resolutions from this conditional distribution. However, due to the information loss during the degradation process, it is difficult to directly infer the pixel-level HR distribution just from a single LR image, especially when the scale factor is large ($\times 8$). While VarSR \cite{VarSR} extracts latent variables in LR space to ease this problem, our method works from another perspective. In order to enhance the expression ability of network, we exploit the non-local self-similarity nature of natural images, which indicates that every patch in a HR image can be well approximated by the sparse representation of atoms in an over-complete dictionary and has been widely utilized in traditional SR methods \cite{2008Image,2014Anchored}. In other words, when the atoms (basis) are fixed, by sampling different coefficients, we can fully explore the diversity of HR patches, thus generate different HR images. 

Suppose HR $x\in \mathbb{R}^{d_x}$, LR $y\in \mathbb{R}^{d_y}$, $d_x = s^2 \times d_y$, where  $s$ denotes the scale factor. We assume that x follows a Gaussian distribution, $x \sim\mathcal N(m+e, \sigma^2I_{d_x})$. As the Figure \ref{fig:bnsr} shows, the deterministic part $m=h(y)$ can be obtained through a deterministic mapping $h$, for example, bicubic up-sampling or any other deterministic SR method like EDSR \cite{enhanced01}, RCAN \cite{2018RCAN}. As for the stochastic part $e \in \mathbb{R}^{s^2 \times d_y}$, we formulate it as the aggregation of small patches $e_i \in \mathbb{R}^{s^2 \times 1} $ with $i \in \{1, 2, \ldots d_y\}$, and each small patch $e_i$ can be represented by coefficients $\omega_i \in \mathbb{R}^{C \times 1}$ under the basis $z\in \mathbb{R}^{s^2 \times C}$, where $e_i = z \cdot \omega_i$, $i \in \{1, 2, \ldots d_y\}$. In order to hold the sparsity of $\omega = (\omega_1, \omega_2, \ldots, \omega_{d_y} )$, we set the gamma prior $\rho = (\rho_1, \rho_2, \ldots, \rho_{d_y})$ for $\omega$, let $\omega_i | \rho_i \sim  \mathcal N(0, \rho_i^{-1})$, $\rho_i \sim \mathcal G(\alpha, \beta)$:
\begin{equation}
p(\omega) = \int p(\omega, \rho)  \,{\rm d} \rho \propto \prod_i {\int \mathcal N(0, \rho_i^{-1})\mathcal G(\alpha, \beta)  \,{\rm d} \rho_i},
\label{01}
\end{equation}
where, $ \alpha $ and $ \beta $ denote the parameters of gamma distribution.

\subsection{Network architecture}\label{sec:arch}

\begin{figure}[t]
	\begin{center}
		\includegraphics[width=1\linewidth]{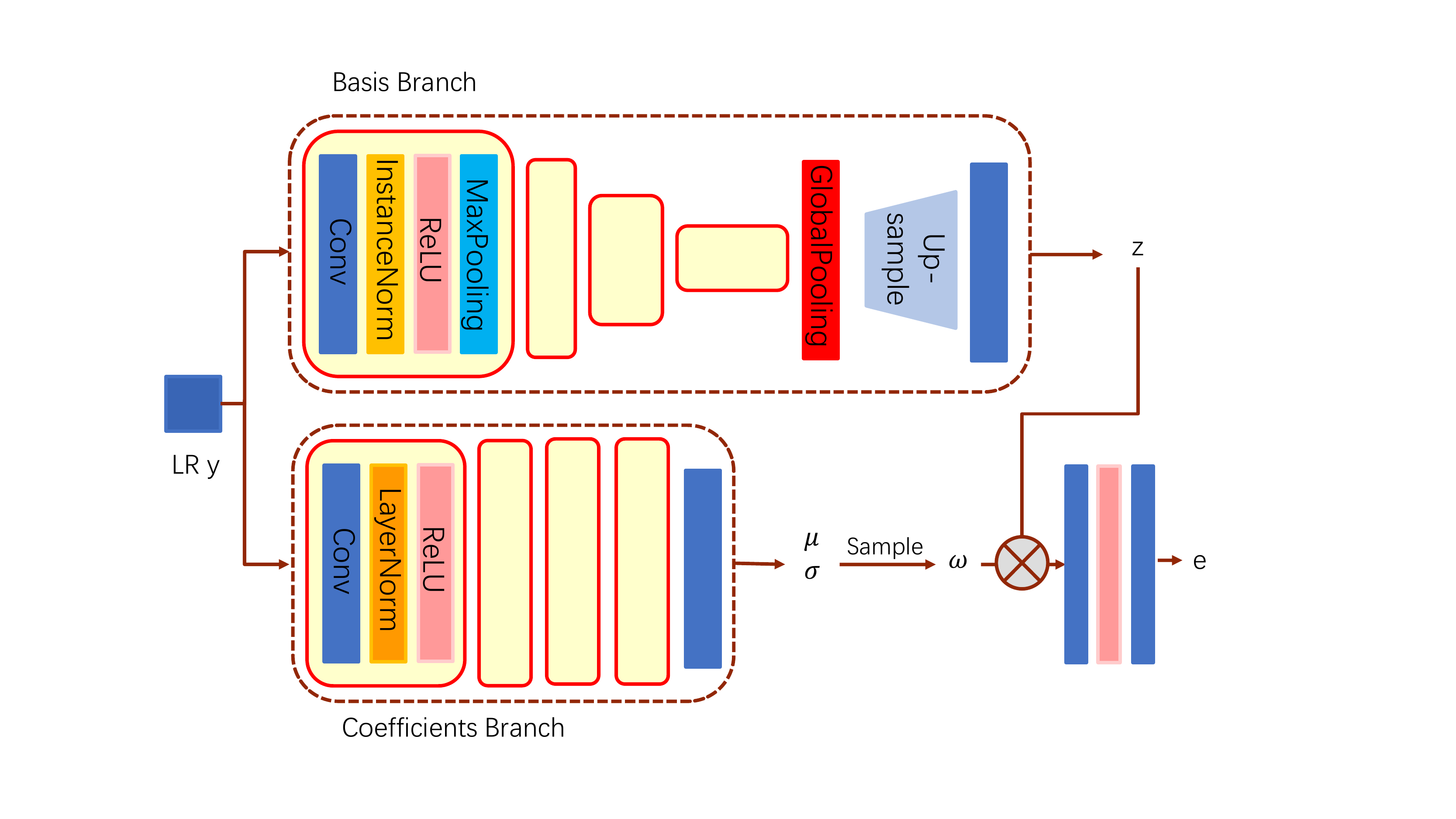}
	\end{center}
	\caption{\textbf{Explorable Module VSpM}. It takes LR $y$ as input, generating (1) basis $z \in \mathbb{R}^{s^2 \times C}$ from the basis branch, (2) distribution parameters $\mu, \sigma \in \mathbb{R}^{d_x \times C}$ of coefficients $\omega$ from the coefficients branch. Finally, we sample coefficients $\omega$ from $\mathcal N(\mu, diag(\sigma^2))$ and obtain stochastic $e$ as output. $\otimes$ denotes matrix multiplication.}
	\label{fig:kbem}
\end{figure}

We designed a variational sparse representation guided explorable module \textbf{VSpM} with two branches, i.e., the basis branch and the coefficients branch, as shown in Figure \ref{fig:kbem}. The basis branch outputs basis $z\in \mathbb{R}^{s^2 \times C}$ , where C denotes the number of basis. The coefficients branch outputs parameters $\mu \in \mathbb{R}^{d_x \times C}$ and $\sigma \in \mathbb{R}^{d_x \times C}$, inferring pixel-wise variational distributions with respect to the sparse coefficients. 

The basis branch is mainly consist of three parts. Firstly, the LR input $y$ goes through $ L $ blocks to generate feature $F_b(y)$:  
\begin{equation}
F_b(y) = F_b^{L}(F_b^{L-1} \dots F_b^{1}(y)).
\label{basis branch}
\end{equation}
And the $F_b^{i}$ in  (\ref{basis branch}) represents the operation of the $i$-th block. Then we use a global pooling to obtain global information $z_0 = F_{gp}(F_b(y)) \in \mathbb{R}^{1^2 \times C}$ of the image.
After that, from the operation of deconvolution, we can upsample  $z_0$ to the size of $s^2 \times C$:
\begin{equation}
z =F_{up}(z_0)= F_{up}(F_{gp}(F_b(y))).
\label{basis branch overall}
\end{equation}

The coefficients branch is simple, which only has a few convolutional layer with layer normalization and ReLU activation. We first infer the parameters $\mu$ and $\sigma$ to estimate the pixel-wise variational distributions with respect to the sparse coefficients. Then we can sample coefficients $\omega$ from the Gaussian distribution $\mathcal N(\mu, diag(\sigma^2))$ at both training and inference stages. Finally, the stochastic part $e$ is restored as the results of matrix multiplication between basis $z$ and coefficients $\omega$. Note that We perform the VSpM in parallel for RGB channels. Also, we adopted the consistency enforcing module (CEM) \cite{BahatM20Explorable} to further enhance the lr-consistency.


\subsection{Training strategies}
We train the network by minimizing
the negative log-likelihood of $x$:
\begin{equation}
L_x = \frac{1}{2\sigma_0^2} {\Vert x - (m+e) \Vert}_2^2.
\label{02}
\end{equation}In order to restrict the distance between the distribution of sampled $\omega$ and the prior distribution as (\ref{01}) shows, we minimize the following KL divergence:
\begin{equation}
L_\omega = \frac{1}{2} \sum_i [{\mu_{\rho i}(\mu_i^2+\sigma_i^2)- \log (\sigma_i^2)+ const}],
\label{03}
\end{equation}
Where, $\mu_{\rho i} = \frac{\alpha + 0.5}{\beta + 0.5 \times (\mu_i^2+\sigma_i^2)}$ represents the $ i $-th elements of the variational parameter $\mu_{\rho}$ with respect to $ \rho $. Finally, we introduce adversarial loss and perceptual loss to enhance the visual quality of SR outputs. Therefore, our total loss function is:
\begin{equation}
L_{total} = L_x + \lambda_{\omega} L_\omega + \lambda_{adv} L_{adv} + \lambda_{per} L_{per}.
\label{04}
\end{equation}
Note that $\sigma_0^2$ in $L_x$ can be absorbed into $\lambda_{\omega}$, $\lambda_{adv}$ and $\lambda_{per}$.

\section{Experiments}\label{sec:4}
In this section, we first showed the datasets and metrics used for training and evaluating VSpSR. Then, we studied the effect of different settings to VSpSR. Finally, we tested the performance of VSpSR on the tasks of SISR $\times 4$ and $\times 8$, and discussed the advantages and limitations of VSpSR.

\begin{table*}[htp]
	\centering
	\caption{Ablation studies on the task of SISR $ \times 4 $. Here, we evaluated all models on the first 20 images selected from the validation dataset of DIV2K. The underline font indicates the optimal settings in each of the sub-studies, while the bold font denotes the best settings across the sub-studies.}
	\label{tab:ablation}
		\begin{tabular}{|c|c|c|c|c|c|c||c|c|c|}
			\hline 
			Model&  \#Basis&  Upsampling&  $ \lambda_{adv} $&  Stochastic $ z $&  Stochastic $ \omega $&  $ \beta $&  LPIPS$\downarrow$&  LR PSNR$\uparrow$&  Div. Score$\uparrow$\\ 
			\hline 
			\#1&  \multirow{3}{*}{256}&  \multirow{3}{*}{Bilinear}&  \multirow{3}{*}{0.01}&  No&  No&  \multirow{3}{*}{0.5}&  0.223&  47.87&  0\\ 
			\#2&  &  &  &  \underline{No}&  \underline{Yes}&  &  0.239&  47.68&  9.161\\
			\#3&  &  &  &  Yes&  Yes&  &  0.309&  48.25&  9.136\\
			\hline
			\#4&  256&  None&  \multirow{2}{*}{0.01}&  \multirow{2}{*}{No}&  \multirow{2}{*}{Yes}&  \multirow{2}{*}{0.5}&  0.320&  47.60&  1.847\\
			\#5&  64&  \underline{Bilinear}&  &  &  &  &  0.308&  47.92&  11.612\\
			\hline
			\#6&  \multirow{3}{*}{\textbf{256}}&  \multirow{3}{*}{Bilinear}&  0.01&  \multirow{3}{*}{No}&  \multirow{3}{*}{Yes}&  \multirow{3}{*}{\textbf{1}}&  0.237&  47.92&  13.251\\
			\#7&  &  &  \textbf{0.1}&  &  &  &  0.280&  47.47&  17.895\\
			\#8&  &  &  1&  &  &  &  0.254&  47.05&  13.325\\
			\hline
			\#9&  256&  Bilinear&  0.1&  No&  Yes&  0.1&  0.220&  47.75&  11.350\\
			\hline 
	\end{tabular}
\end{table*}

\subsection{Dataset and metrics}\label{sub:4.1}
The DIV2K dataset is composed of 800 training images, 100 validation images, and 100 testing images. We will test the performance of our method on the validation dataset since the ground truth of testing dataset is not public. In order to better measure the comprehensive performance of the SR methods, NTIRE 2021 challenge on learning the super-resolution space proposed three metrics to test from three aspects. Before the evaluation, we first generate 10 SR predictions for each LR input in DIV2K validation dataset.

\textbf{LPIPS.} It is very difficult to automatically assess the image perceptual quality. To assess the photo-realism, The challenge will perform a human study on the test set for the final submission. As the SR challenge suggests, in the experiment, we use the Learned Perceptual Image Patch Similarity (LPIPS) \cite{2018LIPIS} distance instead to roughly measure the perceptual quality. 

\textbf{Diversity score.} As mentioned in NTIRE 2021 challenge on learning the super-resolution space \footnote{https://github.com/andreas128/NTIRE21\_Learning\_SR\_Space}, we can use the diversity score to measure the spanning of the SR Space:
\begin{equation}
Div.Score = \frac{LPIPS_{global\_best}-LPIPS_{local\_best}}{LPIPS_{global\_best}},
\label{05}
\end{equation}
where, the local best is obtained by first select \textbf{pixel-wise best} LPIPS score of 10 SR predictions, then compute the average; and the global best is obtained by averaging the whole pixel scores and selecting the best.

\textbf{LR PSNR.} This metric measures the similarity between the SR prediction and the LR image in low-resolution space, which reflects how much the information is preserved during super-resolution. To compute LR PSNR, we should first down-sample the SR predictions and then calculate PSNR. In NTIRE 2021 challenge on learning the super-resolution space, the goal of this metric is to reach 45dB.

\begin{figure}[htp]
	\centering
	\includegraphics[width=0.98\linewidth]{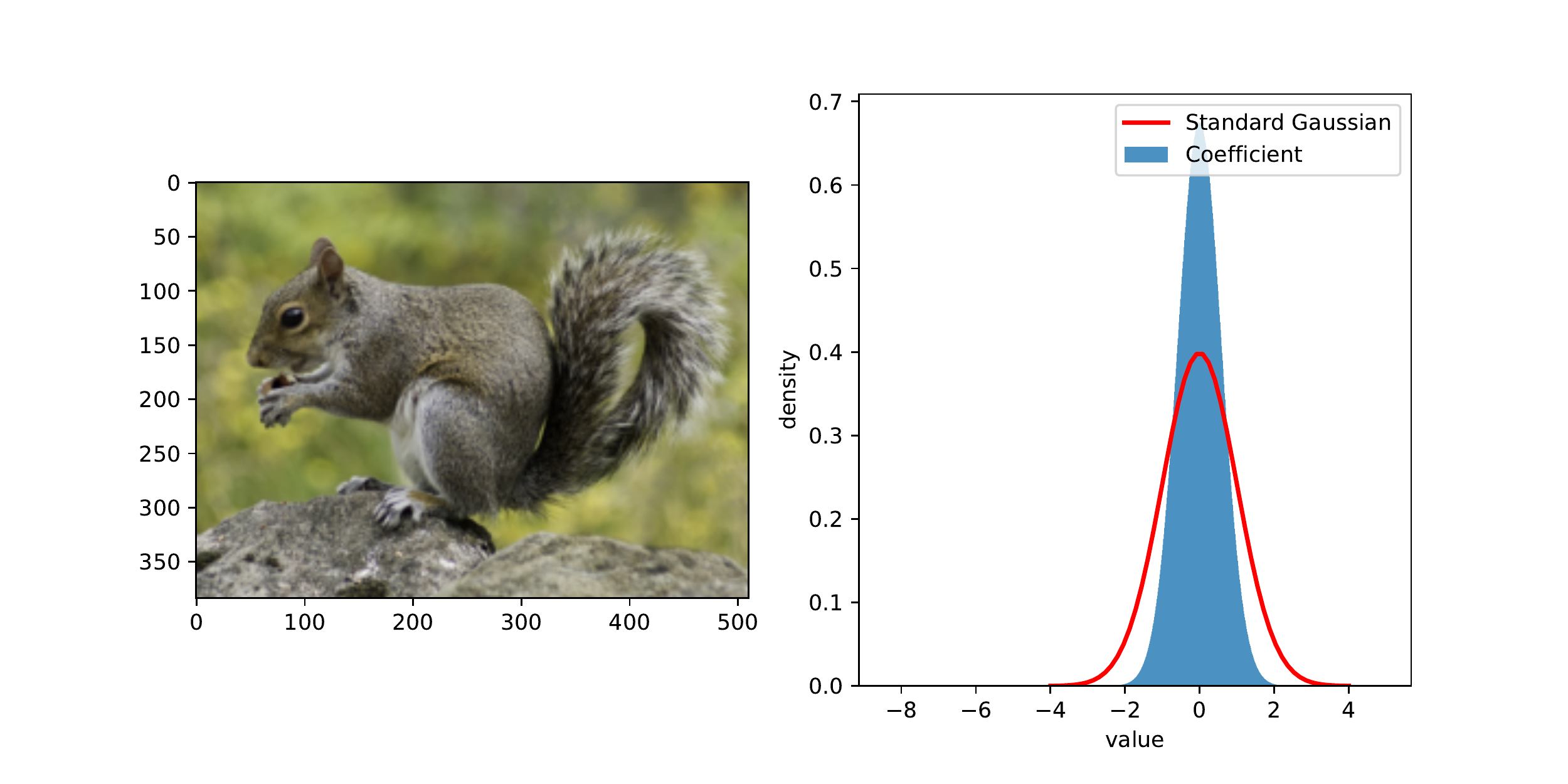}
	\caption{Visualization of the coefficients inferred from a typical LR image, i.e., 0801.png, from the validation dataset of DIV2K. }
	\label{fig:coefficient}
\end{figure}

\subsection{Implementation details}\label{sub:4.2}
There is a data pre-processing before training. To be specific, for $\times 4$ SR tasks, we crop $48 \times 48$ small patches from LR images, and extract corresponding $196 \times 196$ patches from HR images in DIV2K training dataset. For $\times 8$ tasks, we set the LR patch size to $32 \times 32$ and HR patch size to $256 \times 256$. To demonstrate the advantages of the VSpM module, we only use interpolation (bicubic/bilinear) method to generate the deterministic part $ m $ of SR predictions. As for the stochastic part $e$, we set the number of basis to 256, making all the patch-level basis compose a dictionary as over-complete as possible. Besides, the distribution parameters of gamma prior $\rho$ are $\alpha = 3.0$, $\beta = 0.5$. Figure \ref{fig:coefficient} shows that this setting can well restrict the sparsity of coefficients $\omega$.

\begin{figure*}[t]
	\centering
	\includegraphics[width=1\linewidth]{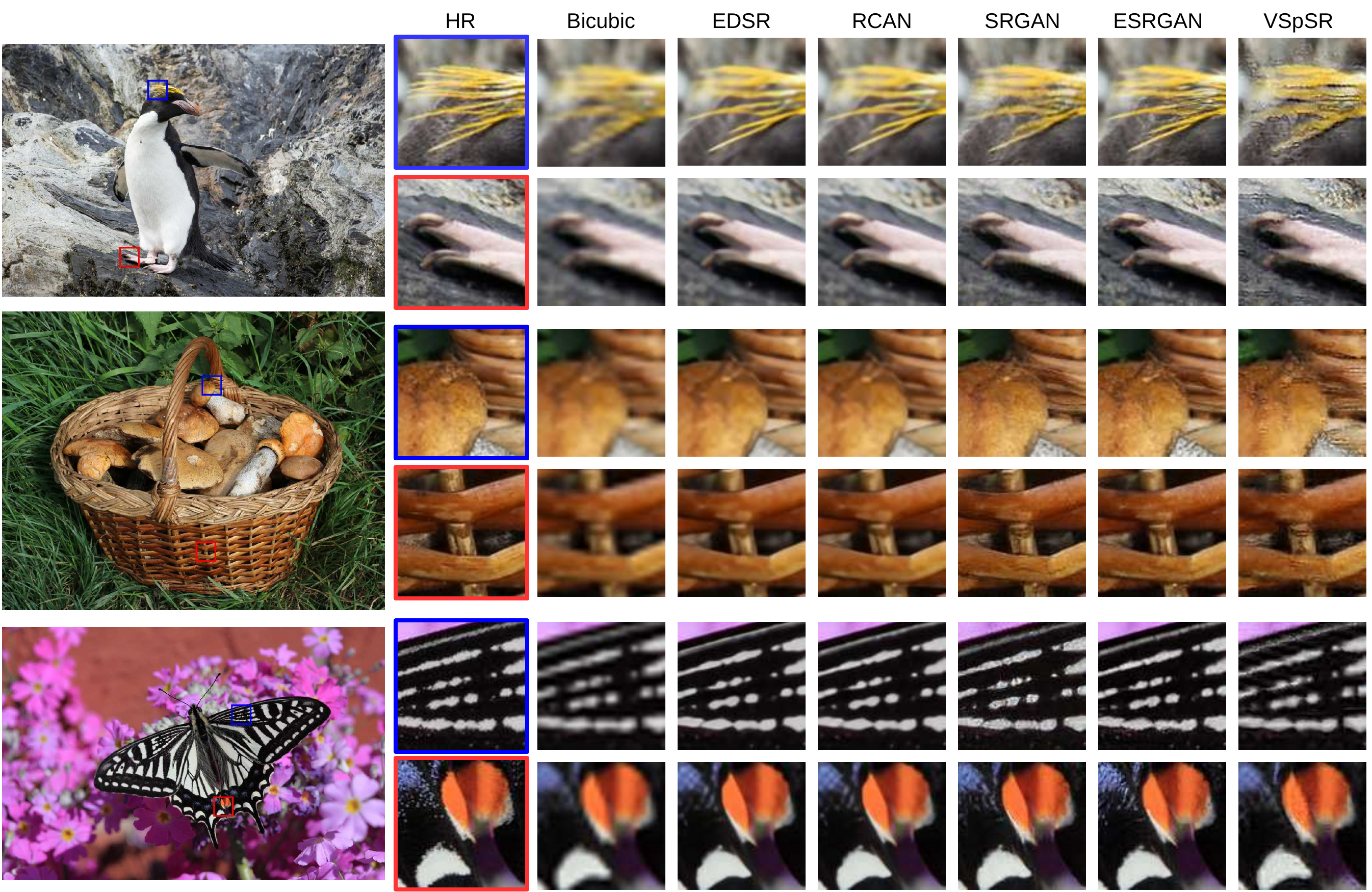}
	\caption{Visualization on the task of SISR $ \times 4 $: three typical examples from DIV2K.}
	\label{fig:sisrx4}
\end{figure*}
\begin{table*}[t]
	\centering
	\caption{Evaluation on the task of SISR $ \times 4 $. Here, the bold value denotes the best performance. \textit{Note that EDSR, RCAN, SRGAN, and ESRGAN were developed to estimate deterministic mappings, and thus their diversity scores are zeros.}}
	\label{tab:sisrx4}
		\begin{tabular}{|c|c|c|c|c|c|c|}
			\hline 
			Method&  Bicubic&  EDSR \cite{enhanced01}&  RCAN \cite{2018RCAN}&  SRGAN \cite{photo01}&  ESRGAN \cite{2018ESRGAN}&  VSpSR\\ 
			\hline 
			LPIPS$ \downarrow $&  0.409&  0.257&  0.254&  0.158&  \textbf{0.115}&  0.277\\ 
			\hline 
			LR PSNR$ \uparrow $&  38.70&  54.11&  \textbf{54.24}&  35.49&  42.61&  47.15\\ 
			\hline
			Div. Score$ \uparrow $&  0&  0&  0&  0&  0&  \textbf{16.120}\\ 
			\hline 
	\end{tabular}
\end{table*}

In training process, all methods are trained by the ADAM optimizer, and settings of parameters are $ \beta_1 = 0.9 $, $ \beta_2 = 0.999 $, and $ \epsilon = 1\times 10^{-8} $. The training of the baseline model is up to 300 epochs, with the initial learning rate of $ 1\times 10^{-4} $, $\lambda_{adv} = 0.01$ and $\lambda_{per} = 0.01$. We let learning rate decreases to 10 percent every 100 epochs. After that, we fine-tune our baseline with settings of $\beta = 1.0$ and $\lambda_{adv} = 0.1$ for extra 100 epochs to improve the performance of models. Finally, we implement our networks with Pytorch and train our models on a device with 40 Intel Xeon 2.20 Ghz CPUs and 4 GTX 1080 Ti GPUs. The whole training process of VSpSR cost about 30 hours on a single GPU. During testing, to evaluate the performance of our method, the metrics mentioned in Section \ref{sub:4.1} are used.

\begin{figure*}[t]
	\centering
	\includegraphics[width=0.9\linewidth, , angle=90]{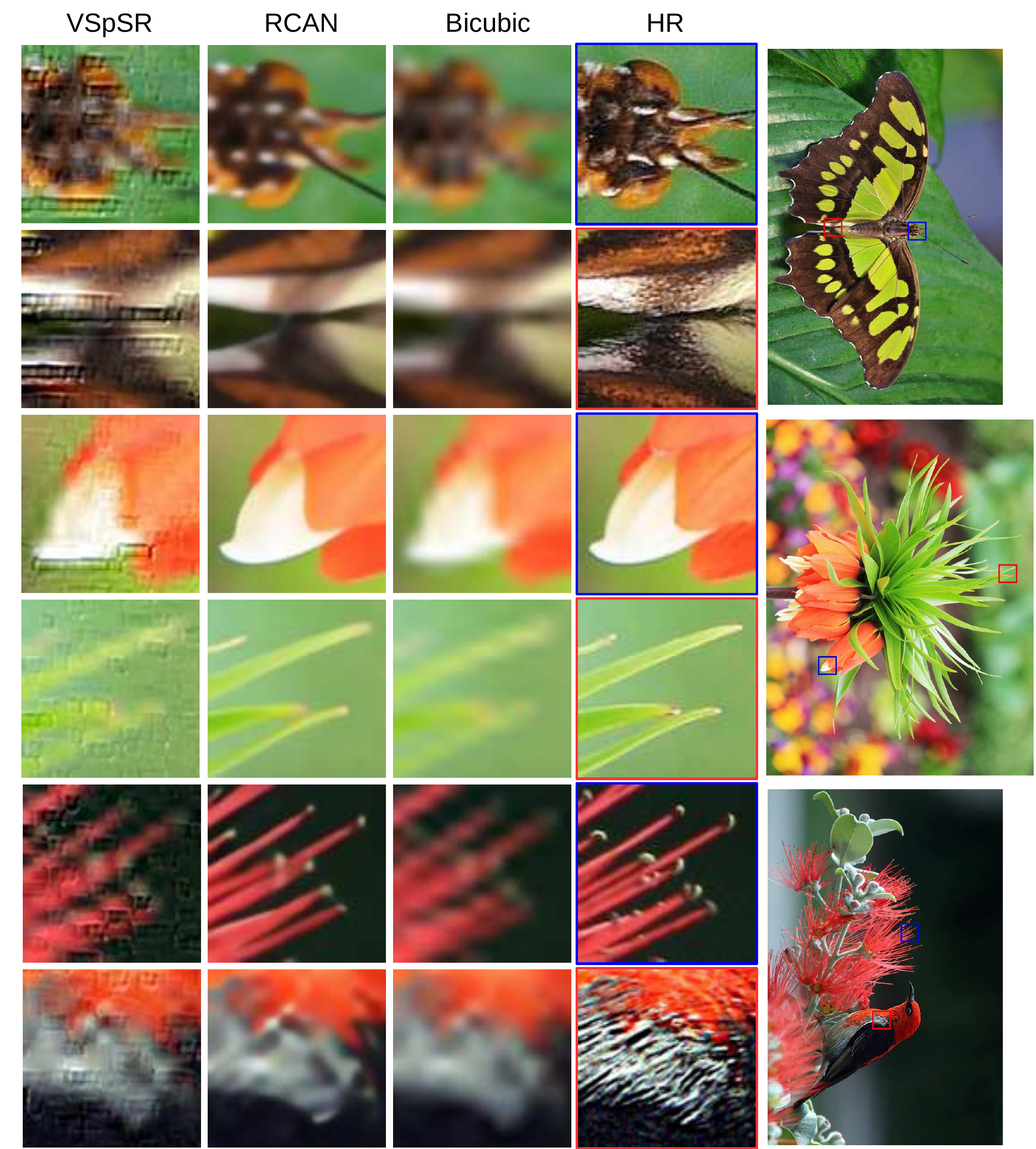}
	\caption{Visualization on the task of SISR $ \times 8 $: three typical examples from DIV2K. Note that VSpSR introduces ``patch-effect'', since we do not explicitly consider the dependency among patches, and that is further discussed in Section \ref{sec:5}.}
	\label{fig:sisrx8}
\end{figure*}

\subsection{Ablation study}\label{sub:ablation}
In this section, we studied the effect of different settings to the performance of VSpSR, including the number of basis (\#Basis), the upsampling manner, the choice of $ \lambda_{adv} $, whether to set the basis $ z $ and the coefficients $ \omega $ to be stochastic, and the selection of $ \beta $. Note that since the evaluation on the whole validation data is computationally expensive, we computed the metrics on the first 20 images of the validation dataset. 

\textbf{Baseline comparison.} We trained 5 models for comparisons (\#1 to \#5), and the first three of which is to study the effect of setting $ z $ and $ \omega $ to be stochastic. The Div. score of model \#1 is zero since the basis and coefficients of VSpSR are deterministic. The comparison between model \#2 and model \#3 shows that setting $ z $ to be stochastic does enlarge the spanning of SR space, but setting $ z $ to be deterministic and $ \omega $ to be stochastic could achieve lower LPIPS value, which results in better Div. score. Thus we adopted such setting in the following studies. 

Besides, we trained models \#4 and \#5 to respectively study the effect of estimating the deterministic $ m $ and that of the \#basis. The comparison between \#2 and \#4 shows that estimating $ m $ could greatly improve the Div. score, this is because that the VSpM module should pay more attention to capture the coarse information from the LR input without the low-frequency information $m$ serves. However, model \#4 tells that the VSpM module alone is still capable of learning SR mapping. Moreover, the comparison between models \#2 and \#5 shows that the larger number of basis could lead to the promotion of the LPIPS, at the cost of increasing the number of total parameters in VSpSR, from 0.27M to 4.4M. 

\textbf{Fine-tuning comparison.} Using the \#2 model as the baseline, we fine-tuned 4 models(\#6 to \#9) for extra 100 epochs to study the effect of $ \lambda_{adv} $ w.r.t. adversarial loss and $ \beta $ w.r.t. Gamma distribution. The comparisons between models \#6, \#7, \#8, and \#9 shows that $ \lambda_{adv}=0.1 $ and $ \beta=1 $ are appropriate choices in terms of Div. score.

\begin{table}[t]
	\centering
	\caption{Evaluation on the task of SISR $ \times 8 $. Here, the bold value denotes the best performance. \textit{Note that RCAN was developed to estimate deterministic mappings, and thus its diversity score is zero.}}
	\label{tab:sisrx8}
		\begin{tabular}{|c|c|c|c|}
			\hline 
			Method&  Bicubic&  RCAN \cite{2018RCAN}&  VSpSR\\ 
			\hline 
			LPIPS$\downarrow$&  0.584&  \textbf{0.404}&  0.508\\ 
			\hline 
			LR PSNR$\uparrow$&  37.16&  \textbf{48.65}&  46.64\\ 
			\hline
			Div. Score$\uparrow$&  0&  0&  \textbf{13.708}\\ 
			\hline 
	\end{tabular}
\end{table}

\begin{figure*}[t]
	\centering
	\includegraphics[width=1\linewidth]{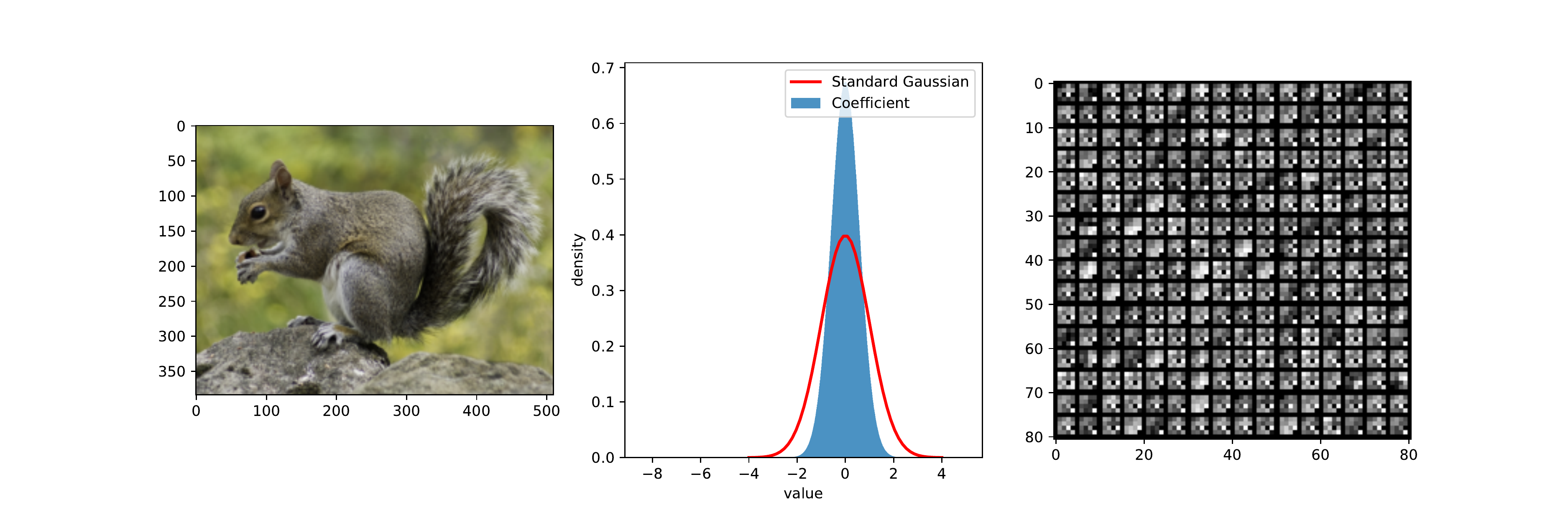}
	\caption{Visualization of a typical example and corresponding sparse representation. Here, the left figure shows a LR image, the middle figure visualizes the distribution of coefficients inferred from the LR image, and the right figure presents the 256 patch-level basis (of size $4\times 4$) estimated from the LR image. }
	\label{fig:discussion}
\end{figure*}

\subsection{Learning SR space $ \times 4 $}\label{sub:5.3}
In this section, we evaluated the performance of VSpSR on the task of SISR $ \times 4 $. Firstly, we adopted the same settings as model \#7 in Table \ref{tab:ablation} for VSpSR, and cropped paired patches of size $ 48\times 48 $ from training images for training. Then, we augmented training patches via flipping and rotation, and minimized the loss function as (\ref{04}) shows to train VSpSR. Finally, we evaluated the performance of VSpSR on the validation dataset of DIV2K via computing LPIPS, LR PSNR, and Div. score, and compare VSpSR with four state-of-the-art SR methods, including two PSNR-oriented, i.e., EDSR and RCAN, and two perceptual-quality-oriented, namely, SRGAN and ESRGAN.

Table \ref{tab:sisrx4} shows the quantitative results of compared methods. Since RCAN is PSNR-oriented while ESRGAN is perceptual-quality-oriented, they achieve the best LR PSNR and LPIPS among all methods, respectively. However, EDSR \cite{enhanced01}, RCAN \cite{2018RCAN}, SRGAN \cite{photo01}, and ESRGAN \cite{2018ESRGAN} are deterministic models, and their diversity score are zeros. Being different from these methods, our VSpSR could generate diverse SR image from a single LR image, since we made the coefficients $ \omega $ to be stochastic by the variational sparse representation. To qualitatively evaluate the performance of VSpSR, we visualized three typical examples in Figure \ref{fig:sisrx4}. This figure shows that the perceptual-quality-oriented methods, i.e., SRGAN and ESRGAN could generate more details, which is consistent with the quantitative results.

\subsection{Learning SR space $ \times 8 $}
In this section, we evaluated the performance of VSpSR on the task of SISR $ \times 8 $. Firstly, we adopted the same settings as model \#7 in Table \ref{tab:ablation} for VSpSR, and cropped paired patches of size $ 32\times 32 $ from training images. Then, we augmented training patches via flipping and rotation, and minimized the loss function as (\ref{04}) shows to train VSpSR. Finally, we evaluated the performance of VSpSR on the validation dataset of DIV2K via computing LPIPS, LR PSNR, and Div. score. Since RCAN \cite{2018RCAN} released the model for SISR $ \times 8 $, we compared it with our VSpSR.

Table \ref{tab:sisrx8} shows the quantitative results of compared methods, including bicubic, RCAN, and our VSpSR. Although RCAN could achieve the best LPIPS and LR PSNR, its diversity score is zero since RCAN is a deterministic model. Being different from RCAN, our VSpSR could reconstruct diverse SR images, since the coefficients of proposed variational sparse representation are stochastic. To qualitatively evaluate the performance of VSpSR, we visualized three typical examples from the validation dataset in Figure \ref{fig:sisrx8}. This figure shows that RCAN could generate higher quality images than VSpSR. Besides, VSpSR could introduce ``patch-effect'', since we do not explicitly consider the dependency among patches, and that will be further discussed in Section \ref{sec:5}. Although VSpSR cannot perform robustly as RCAN in reconstructing details, it has the advantage of generating diverse SR images which are consistent with a single LR image, and that is one of the keys of learning SR space.

\section{Discussion}\label{sec:5}
The advantage of VSpSR is that it could greatly expand SR space compared with the deterministic models, but ``patch-effect'' is introduced due to the patch-level sparse representation. Concretely, the conventional sparse representation is aimed at building an dictionary, such that each small patch could be sparsely represented by the dictionary. The expanded space with respect to sparse representation is determined by the dictionary, and therefore an over-complete dictionary is required. However, such representation is deterministic and computationally expensive, and thus cannot be applied to learn SR space. To tackle the difficulty, we proposed the variational sparse representation framework, i.e., VSpSR, whose coefficients follow a sparse prior and could be repeatedly sampled from a variational distribution. To further understand VSpSR, we shows the distribution of coefficients and visualizes the basis inferred from a typical LR image in Figure \ref{fig:discussion}. For VSpSR, the basis determines the expanded SR space, while a sample of sparse coefficients is corresponding to a SR image in the space. That means increasing the number of basis could rise the diversity of SR space, but that would also increase the computational complexity. Therefore, exploring more efficiency methods of increasing the diversity of SR space is required. Besides, we only study the patch-wise sparse representation, and do not explicitly model the dependency among patches. That would introduce ``patch-effect'' as Figure \ref{fig:sisrx8} shows for big scaling factors, e.g., $ \times 8 $. In reality, different patches may be highly similar, and thus explicitly modeling such dependency is appealing.

\begin{table}[htp]
	\centering
	\caption{Preliminary results $ \times 4 $. The bold font denotes our results.}
	\label{tab:resultsx4}
	\begin{tabular}{llll}
		\toprule
		{} &  LPIPS & LR PSNR & Div. Score \\
		Team            &        &         &            \\
		\midrule
		svnit\_ntnu      &  0.355 &   27.52 &      1.871 \\
		SYSU-FVL        &  0.244 &   49.33 &      8.735 \\
		nanbeihuishi    &  0.161 &   50.46 &     12.447 \\
		SSS             &  0.110 &   44.70 &     13.285 \\
		\textbf{Ours}     &  \textbf{0.273} &   \textbf{47.20} &     \textbf{16.450} \\
		FutureReference &  0.165 &   37.51 &     19.636 \\
		SR\_DL           &  0.234 &   39.80 &     20.508 \\
		CIPLAB          &  0.121 &   50.70 &     23.091 \\
		BeWater         &  0.137 &   49.59 &     23.948 \\
		Deepest         &  0.117 &   50.54 &     26.041 \\
		njtech\&seu      &  0.149 &   46.74 &     26.924 \\
		\bottomrule
	\end{tabular}
\end{table}

\begin{table}[htp]
	\centering
	\caption{Preliminary results $ \times 8 $. The bold font denotes our results.}
	\label{tab:resultsx8}
	\begin{tabular}{llll}
		\toprule
		{} &  LPIPS & LR PSNR & Div. Score \\
		Team            &        &         &            \\
		\midrule
		svnit\_ntnu      &  0.481 &   25.55 &      4.516 \\
		SYSU-FVL        &  0.415 &   47.27 &      8.778 \\
		SSS             &  0.237 &   37.43 &     13.548 \\
		\textbf{Ours}     &  \textbf{0.496} &   \textbf{46.78} &     \textbf{14.287} \\
		SR\_DL           &  0.311 &   42.28 &     14.817 \\
		FutureReference &  0.291 &   36.51 &     17.985 \\
		CIPLAB          &  0.266 &   50.86 &     23.320 \\
		BeWater         &  0.297 &   49.63 &     23.700 \\
		Deepest         &  0.259 &   48.64 &     26.941 \\
		njtech\&seu      &  0.366 &   29.65 &     28.193 \\
		\bottomrule
	\end{tabular}
\end{table}

\section{Conclusion}\label{sec:6}
The NTIRE 2021 challenge on learning the super-resolution space is difficult since inference of SR space instead of single SR prediction increases the amount of details to restore from a single LR input. Besides this, it is more difficult to hold the balance between the spanning of SR space and the consistency in LR space, when promoting visual quality as much as possible. To tackle these difficulties, we have proposed a variational sparse framework implemented via neural network to solve the SR challenge raised in NTIRE 2021. Specifically, we design a two-branch module, i.e., VSpM,  to explore the SR space. The basis branch of VSpM extracts patch-level basis from the LR input, and the coefficients branch infers pixel-wise variational distributions with respect to the sparse coefficients. Therefore, we could obtain different sparse representations by repeatedly sampling coefficients, and thus generate diverse HR images. Finally, we have tested the performance of VSpSR in Section \ref{sec:4} to show its effectiveness in conducting explorable super-resolution, and discussed the advantages and limitations of VSpSR in Section \ref{sec:5}. According to the preliminary results as Tables \ref{tab:resultsx4} and \ref{tab:resultsx8} show, our team ranks 7-th in terms of released Div. scores \cite{lugmayr2021ntire}.

\textbf{Acknowledgement.} This work was funded by the National Natural Science Foundation of China (grant no. 61971142 and
62011540404) and the development fund for Shanghai talents (no.2020015).

{\small
	\bibliographystyle{ieee_fullname}
	\bibliography{egbib}
}
	
\end{document}